\documentclass[letterpaper, 10 pt, conference]{ieeeconf}
\IEEEoverridecommandlockouts
\usepackage[T1]{fontenc}
\usepackage[utf8]{inputenc}
\usepackage{cite}
\usepackage{amsmath,amssymb,amsfonts}
\usepackage{algorithmic}
\usepackage{graphicx}
\usepackage{textcomp}
\usepackage{xcolor}
\usepackage{threeparttable}
\usepackage{booktabs}
\usepackage{multirow}
\usepackage{stfloats}
\usepackage{pifont}
\usepackage{orcidlink}

\def\BibTeX{{\rm B\kern-.05em{\sc i\kern-.025em b}\kern-.08em
    T\kern-.1667em\lower.7ex\hbox{E}\kern-.125emX}}

\begin{document}
\bstctlcite{IEEEexample:BSTcontrol} 

\title{\LARGE \bf REACT: A Fully Spiking State-Space Model for\\
Real-Time Event-Driven Temporal Perception}

\author{
    Geoffroy Keime\\
    CerCo, CNRS UMR5549 \\
    Université de Toulouse \\
    Toulouse, France \\
    IPAL, CNRS IRL, Singapore \\
    geoffroy.keime@cnrs.fr \orcidlink{0009-0000-3451-9458}
    \and
    Nicolas Cuperlier \\
    ETIS, CY Cergy Paris Université \\
    CNRS, ENSEA \\
    Cergy, France \\
    IPAL, CNRS IRL, Singapore \\
    nicolas.cuperlier@cyu.fr \orcidlink{0000-0001-7610-3143}
    \and
    Benoit R. Cottereau \\
    CerCo, CNRS UMR5549 \\
    Université de Toulouse \\
    Toulouse, France \\
    IPAL, CNRS IRL, Singapore \\
    benoit.cottereau@cnrs.fr \orcidlink{0000-0002-2624-7680}
    \thanks{This work was supported by the French Defense Innovation Agency (AID) under grant number 2023 65 0082.}
}

\maketitle
\thispagestyle{empty}
\pagestyle{empty}

\begin{abstract}
Robotic systems operating in dynamic environments require visual perception that evolves continuously with the incoming sensory stream. Event cameras provide microsecond temporal resolution and asynchronous sensing, but most learning-based methods accumulate events into frames or temporal bins, introducing an integration delay that can limit fast reaction. Here we propose REACT, a fully spiking state-space model for event-driven temporal perception that processes raw events one by one, without temporal accumulation. REACT uses a complex-valued spiking neuron, C-SiLIF, whose continuous-time dynamics are driven by the physical inter-event interval, allowing its internal state to evolve at the temporal resolution of individual events. We evaluate REACT on gesture recognition and time-to-collision (TTC) estimation from full-field event streams, without a target bounding box or localization input. On EvTTC, REACT achieves a 9.59\% relative TTC error with 4.6 ms end-to-end inference latency, within 0.15 percentage points of the best learned method while requiring no target prior. At the dataset's mean approach speed, this latency corresponds to only 4 cm of vehicle motion, compared with 1 m for the fastest competing learned method. REACT further supports anytime TTC prediction, zero-shot transfer to a different driving sequence, and INT8 quantization, reducing the estimated energy consumption from 18.5 to 2.8 mJ per 32,768 events. These results show that event-driven spiking state-space dynamics can provide low-latency, continuously updated temporal perception for reactive robotic systems.
\end{abstract}


\section{Introduction}
\label{introduction}

Robotic systems operating in dynamic environments require visual perception that evolves continuously with the incoming sensory stream \cite{gallegoReview2022}. Event cameras provide microsecond temporal resolution and asynchronous sensing, making them particularly attractive for low-latency, asynchronous event-driven robotic perception. However, most learned event-based methods accumulate events into frames or temporal bins before inference, introducing a temporal integration delay and an artificial discretization of the sensor stream \cite{gallegoReview2022, gehrigE2E2019, lagorceHOTSHierarchyEventBased2017f}. Such aggregation prevents the model from updating its state at the native event timescale and can limit timely perception in fast-moving environments.

Spiking neural networks (SNNs) naturally process event streams through sparse, asynchronous updates, while state-space models (SSMs) provide an efficient framework for modeling temporal dependencies through compact dynamical states. Recent works have combined these paradigms for event-based vision, including spiking formulations of SSMs and SSM-inspired spiking neurons, but they generally do not combine event-driven spiking computation with continuous-time state evolution directly at the irregular timestamps of incoming events. As a result, the temporal structure of the event stream remains partially constrained by discrete processing or temporal aggregation.

We propose REACT (Real-time Event-driven Asynchronous Computation for Temporal perception), a fully spiking, asynchronous event-driven state-space model that processes raw events sequentially as they arrive. REACT uses the complex-valued spiking neuron C-SiLIF, whose continuous-time dynamics are driven by the measured inter-event interval. The internal state is therefore updated event by event, with the elapsed time between consecutive events explicitly incorporated into the dynamics, without requiring a fixed temporal window or event accumulation. This enables REACT to preserve fine-grained temporal information while retaining the efficient linear recurrence of an SSM.

We evaluate REACT on two complementary temporal perception tasks: gesture recognition on DVS128 Gesture and time-to-collision (TTC) estimation on EvTTC. For TTC estimation, REACT processes the full field of view without a supplied target bounding box, requiring temporal collision-relevant information to emerge directly from the asynchronous event stream. This provides a direct test of asynchronous event-driven temporal perception in a dynamic driving scenario. We further investigate zero-shot transfer, anytime prediction, and INT8 quantization, assessing both the temporal flexibility and practical efficiency of the proposed approach.

\begin{figure*}[t]
    \centering
    \includegraphics[width=\textwidth]{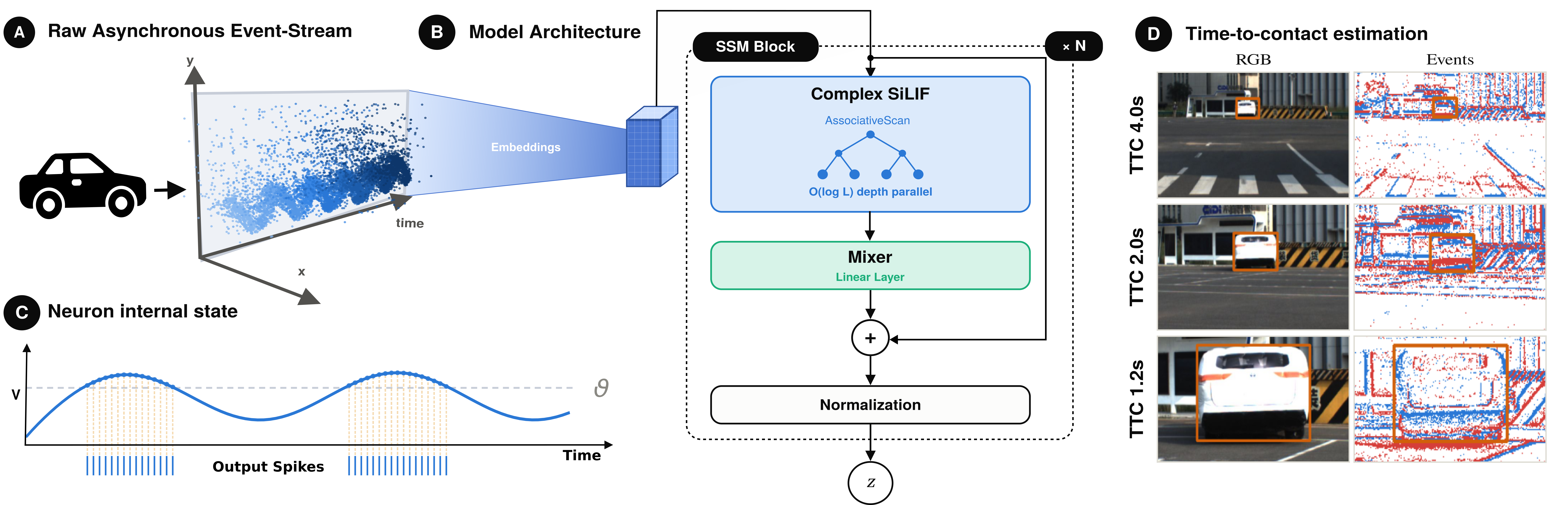}
\caption{\textbf{Overview of REACT.}
(\textbf{A}) Vehicle motion generates an asynchronous stream of events.
(\textbf{B}) Each event is embedded into a fixed-dimensional vector and processed sequentially by spiking state-space blocks, whose temporal dynamics are driven by the physical inter-event interval.
(\textbf{C}) The C-SiLIF neurons emit binary spikes when their membrane potential crosses the threshold, without resetting the underlying state, thereby preserving continuous-time state dynamics. 
(\textbf{D}) Examples of event-based visual inputs at 4, 2, and 1.2~s before collision. Red and blue spikes correspond to \textit{On} and \textit{Off} events, respectively. The associated RGB frames are shown for illustration only and were not used by the model.}
    \label{fig:pipeline}
\end{figure*}

The main contributions of this work are as follows:





\begin{itemize}
    \item We introduce REACT, a new fully spiking state-space architecture for raw event streams, combining asynchronous event-by-event processing with continuous-time state evolution driven by the physical inter-event interval. This eliminates temporal accumulation while preserving continuous-time temporal dynamics at the native timescale of the sensor.
    \item We demonstrate target-free, anytime time-to-contact (TTC) estimation directly from full-field event streams, without temporal accumulation or a supplied target bounding box. REACT continuously updates its prediction as events arrive and achieves 4.6 ms end-to-end latency with a 9.59\% relative TTC error on EvTTC.
    \item We show that the learned temporal representation supports both anytime prediction and zero-shot transfer across datasets, demonstrating its ability to provide useful temporal estimates under varying observation horizons and sensing conditions.
    \item We quantify the efficiency potential of the fully spiking formulation for neuromorphic deployment. REACT maintains competitive gesture-recognition accuracy with a substantially smaller state-space component, while INT8 quantization reduces the estimated TTC energy from 18.5 to 2.8 mJ per 32,768 events.
\end{itemize}

Code will be made publicly available upon publication.

\section{Related Works} \label{related_works} 
\subsection{State Space Models and Spiking Neural Networks} \label{rel_ssm} 
Most current event-based vision models rely on an intermediate event representation, aggregating events into pseudo-frames through hand-designed or learned strategies. While effective, these representations discard part of the fine-grained timing provided by the sensor. Processing events directly preserves this temporal resolution, but requires architectures capable of efficiently modeling long asynchronous sequences. State space models (SSMs) have emerged as efficient sequence models, scaling linearly with sequence length while maintaining bounded computational and memory costs \cite{guEfficientlyModelingLong2022, smithSimplifiedStateSpace2023, guMambaLinearTimeSequence2024}. 
The linear ordinary differential equations underlying continuous-time SSMs are structurally analogous to the subthreshold dynamics of leaky integrate-and-fire (LIF) neurons. This correspondence has motivated both spiking formulations of SSMs \cite{balRethinkingSpikingNeural2024, stanLearningLongSequences2023} and the derivation of neuron models from SSM parameterizations \cite{fabreSiLIFStructuredState2026}, including SSM-inspired adaptive LIF \cite{bittarSurrogateGradientSpiking2022, baronigAdvancingSpatioTemporalProcessing2024a} and resonate-and-fire neurons \cite{IZHIKEVICH2001883, higuchi2024balanced}. SSMs have  also been applied directly to raw event streams, avoiding frame-based representations \cite{schoneScalableEventbyeventProcessing2024, zubicStateSpaceModels2024, zhangComputeinmemoryImplementationState2026}. These approaches efficiently extract spatio-temporal features from asynchronous data and offer pathways toward compute-in-memory implementations \cite{zhangComputeinmemoryImplementationState2026}. However, their computation remains fundamentally dense, requiring state updates across the model as events are processed. They therefore do not fully exploit the sparse event-driven activity that underpins the energy efficiency of neuromorphic computing. This limitation is particularly relevant for temporal prediction tasks in which low latency and fine-grained timing are critical, such as time-to-contact estimation.
\subsection{Time-to-contact Estimation} \label{rel_ttc} 
Time-to-contact (TTC) denotes the remaining time before an object reaches the observer under its current relative motion. Unlike metric depth, TTC does not require recovering absolute scale or metric scene structure, making it a compact cue for reactive behaviors such as braking, landing, and obstacle avoidance \cite{cladyAsynchronousVisualEventbased2014}. Because informative visual changes become increasingly concentrated near the collision endpoint, latency is critical for accurate real-time TTC prediction \cite{falanga2020}. Event cameras are therefore particularly well suited to TTC estimation through their asynchronous response, high temporal resolution, and high dynamic range. Event-based TTC estimation has traditionally relied on model-based approaches that fit explicit geometric or kinematic models to event streams \cite{cladyAsynchronousVisualEventbased2014, gallego2018unifying, li2024eventaided, mcleod2022globally, dinaux2021faith, nunes2023ttcmap}. These methods typically recover the focus of expansion, motion parameters, or scene depth before deriving TTC. A smaller body of work preserves sensor asynchrony by updating a filter event by event \cite{wang2024asynchronous}, but still relies on a hand-specified observation model. Their performance is consequently constrained by assumptions such as prescribed motion patterns or planar and rigid-translation models, as well as by iterative inference \cite{li2026eap}. Learning-based TTC estimation remains comparatively underexplored, with only a few recent approaches \cite{bisulco2025evttc, li2026eap}. Garl-TTC \cite{li2026eap} is currently the only dedicated object-level framework. To the best of our knowledge, the combination of fully spiking, event-by-event processing and explicit state-space temporal dynamics remains largely unexplored for TTC estimation.

\section{Methods}
\label{methods}

\subsection{Event stream and input encoding}
\label{sec:input}

An event camera outputs an asynchronous stream of $L$ tuples
$\mathcal{E} = \{(t_k, x_k, y_k, p_k)\}_{k=1}^{L}$, where $t_k$ is the
timestamp, $(x_k,y_k)$ the pixel coordinate on an $H \times W$ sensor, and
$p_k \in \{0,1\}$ the polarity of the brightness change. Most learning-based
pipelines first accumulate this stream into frames, voxel grids, or fixed time
bins. REACT consumes it directly: no event representation is constructed at any
stage, and the model state is updated once per event. Each event is reduced to a spatial channel index:
\begin{equation}
j_k = p_k \cdot H W + y_k \cdot W + x_k \;\in\; \{0, \ldots, 2HW-1\},
\label{eq:index}
\end{equation}
and a physical inter-event interval $\Delta t_k = t_k - t_{k-1}$, with
$\Delta t_1 = 0$ by convention, intervals are expressed in milliseconds. The stream is thus a pair of aligned sequences
$(j_k, \Delta t_k)$: $j_k$ carries \emph{where} the event occurred and
$\Delta t_k$ carries \emph{when}, the latter entering the network as a
continuous quantity rather than as a bin index. The channel index is mapped to
a $D$-dimensional embedding $\mathbf{z}_k = \mathrm{Embed}(j_k) \in
\mathbb{R}^{D}$ followed by layer normalisation, through a learned lookup table
$\mathbf{E} \in \mathbb{R}^{2HW \times D}$ that can be read as a per-pixel,
per-polarity synaptic fan-in. Variable-length streams are right-padded and
masked for batched training; padded positions contribute to neither the loss nor
the reported firing rates. Fig.~\ref{fig:pipeline} summarises the resulting
pipeline.

\subsection{The C-SiLIF neuron}
\label{sec:csilif}

The core computational unit of our SNN is the complex-valued C-SiLIF neuron introduced
by~\cite{fabreSiLIFStructuredState2026}, whose subthreshold dynamics form a diagonal linear time-invariant state-space model with one complex state per channel, a log-parameterized decay, a learnable timestep, and S4D-Lin initialization~\cite{guParameterizationInitializationDiagonal2022}. Fabre et
al.\ evaluate it on fixed-bin audio streams. We retain their formulation and
drive the discretisation with the physical inter-event interval $\Delta t_k$
rather than a fixed step, which is what makes event-by-event operation possible.
Each of the $D$ channels carries one independent state $u \in \mathbb{C}$,
evolving as:
\begin{equation}
\frac{du}{dt} = \Lambda u + b\, x(t), \quad
\Lambda = -e^{\lambda_{\mathrm{re}}} + i\,\lambda_{\mathrm{im}}, \quad
b = b_{\mathrm{re}} + i\,b_{\mathrm{im}},
\label{eq:ct}
\end{equation}
where $\lambda_{\mathrm{re}}$ is a log-parametrized decay rate,
$\lambda_{\mathrm{im}}$ an angular frequency, and $b$ a complex input gain,
all learned per neuron.

Discretisation rescales the measured inter-event interval by a per-neuron learnable factor $e^{\rho}$, giving an effective step $\delta_k = e^{\rho}\,\Delta t_k$, so each neuron measures time in units of its own, and event-driven update.
\begin{align}
\alpha_k &= \exp\!\big( (-e^{\lambda_{\mathrm{re}}} + i\,\lambda_{\mathrm{im}})\, \delta_k \big), \label{eq:alpha}\\
u_k &= \alpha_k\, u_{k-1} + b\, \mathbf{z}_k, \label{eq:state}\\
s_k &= \Theta\big( 2\,\mathrm{Re}(u_k) - \vartheta \big), \label{eq:spike}
\end{align}
with $\Theta$ the Heaviside step function, $\vartheta$ a learnable firing
threshold, and $s_k \in \{0,1\}$ the emitted spike. Events enter as Dirac impulses at their timestamps, so the physical inter-event interval enters the recurrence only through the transition coefficient $\alpha_k$. The factor of two follows from the real-valued observation of the conjugate-symmetric state pair, since $u+\bar u=2\operatorname{Re}(u)$. 

Importantly, no reset is applied to $u_k$ after a spike. The state recurrence therefore remains linear, while the spiking nonlinearity acts only as a readout. Because the state transition coefficient $\alpha_k$ depends on the physical inter-event interval, the temporal dynamics are driven directly by the timing of incoming events. The dynamics are also stable by construction:
$e^{\lambda_{\mathrm{re}}}>0$ implies $|\alpha_k|<1$ for any positive interval. Equivalently, each neuron acts as a damped harmonic resonator with impulse response
$|b|\,e^{-e^{\lambda_{\mathrm{re}}}t}\cos(\lambda_{\mathrm{im}}t+\varphi)$.
The phase $\varphi$ therefore controls the temporal phase of the neuron's oscillatory impulse response, enabling different neurons to preferentially respond to different temporal patterns without introducing an explicit delay parameter. Gradients are propagated through
\eqref{eq:spike} with an arctangent surrogate of width $2$.

Following the S4D-Lin initialisation adopted for C-SiLIF in ~\cite{fabreSiLIFStructuredState2026}, we set
$e^{\lambda_{\mathrm{re}}}=0.5$ and $\lambda_{\mathrm{im}}=\pi /s$ as shared
initial values, sample $e^{\rho}$ log-uniformly over
$10^{-3}$--$10^{-1}$, and initialise the input gain as $b_{\mathrm{re}}\sim\mathcal{U}(0,1)$ and $b_{\mathrm{im}}=0$. The parameters $\lambda_{\mathrm{re}}$ and $\lambda_{\mathrm{im}}$ define the initial pole location, while the heterogeneous timestep $\rho$ provides
each neuron with a distinct temporal scale, and therefore a distinct effective decay and oscillation frequency. This diversity avoids starting
all neurons from a homogeneous temporal configuration and provides a broad set of initial temporal responses for optimisation.

\subsection{Sequential inference, parallel training}
\label{sec:scan}

Equation~\eqref{eq:state} is a first-order linear recurrence
$u_k = \alpha_k u_{k-1} + b_k$ whose transitions compose associatively,
\begin{equation}
(A_j,B_j) \circ (A_i,B_i) = (A_j A_i,\; A_j B_i + B_j),
\label{eq:assoc}
\end{equation}
so the state sequence $\{u_k\}_{k=1}^{L}$ can be obtained by an inclusive
Hillis--Steele prefix scan in $O(\log L)$ sequential depth and $O(L)$ total
work, parallelised across batch elements and channels. We use this scan during
training and batched offline evaluation only. At deployment the model runs
strictly sequentially, one state update per incoming event, which is the regime
assumed by the latency and energy figures of Secs.~\ref{sec:ttc}
and~\ref{res_energy}.

\subsection{Network architecture}
\label{sec:arch}

The network body is a stack of $N = 2$ residual blocks with $D = 128$ channels,
operating on the token sequence produced by the embedding. Each block applies
C-SiLIF independently along the temporal dimension of every channel, updating
its state according to the inter-event intervals and emitting a binary spike
train, and a channel-mixing linear stage then integrates information across
features. The block output is combined with its input through a residual
connection, which keeps optimisation stable as blocks are stacked. Every signal
crossing a block boundary is therefore a spike train, and each C-SiLIF layer is
driven by spikes throughout the network, so all synaptic operations reduce to accumulations.

Because sequence length grows with the number of incoming events, consecutive
tokens are merged between blocks by masked-mean pooling, with their inter-event
intervals summed so that the compressed sequence still represents the correct
elapsed physical time. We pool by a factor of $32$ on DVS128-Gesture and apply
no pooling on EvTTC, which is the source of the higher per-event energy reported
in Sec.~\ref{res_energy}.

After the last block, tokens are masked-mean-pooled over valid positions and
mapped by a single linear layer to the task output. On neuromorphic substrates
this readout corresponds to a non-firing integrator, whose membrane potential is
read out directly.

\subsection{Task formulation and metrics}
\label{sec:task}

For gesture recognition the readout produces a distribution over the $11$
classes, trained with cross-entropy and label smoothing.

For time-to-contact we regress the inverse contact time $\eta = 1/\mathrm{TTC}$
rather than the time itself, since $\mathrm{TTC}$ diverges under slow approach
while $\eta$ remains bounded. The readout emits a mean $\mu_i$ and a
log-variance $s_i$ in $\log\eta$ space, trained with a Gaussian negative
log-likelihood,
\begin{equation}
  \mathcal{L}_{\mathrm{NLL}} = \frac{1}{N}\sum_{i=1}^{N}
  \Big[\tfrac{1}{2}e^{-s_i}\big(\mu_i - \log\eta_i\big)^2 + \tfrac{1}{2}s_i\Big],
  \label{eq:nll}
\end{equation}
so that $\hat\eta = e^{\mu}$ and $\widehat{\mathrm{TTC}} = e^{-\mu}$, with
$s_i$ providing a per-estimate uncertainty alongside the prediction.

We report the EvTTC benchmark metric~\cite{sun2025evttc},
\begin{equation}
  e_{\mathrm{TTC}} = \frac{|\widehat{\mathrm{TTC}} - \mathrm{TTC}|}{\mathrm{TTC}}
  \times 100\% ,
  \label{eq:ettc}
\end{equation}
averaged per sequence over its full range, so that every row of
Table~\ref{tab:ttc-related-work} is scored identically. Because
\eqref{eq:ettc} normalises by ground-truth time, a fixed absolute error counts
for little when the target is far and heavily when it is close, and a sequence
average is dominated by its far portion. We therefore also report a
criticality-weighted log-ratio error, $\mathrm{MiD}_w$, which averages
$|\log \hat\eta - \log \eta|$ within ground-truth TTC bands and reweights them
towards imminent contact, with weights $0.5$, $0.3$ and $0.1$ on $[0,3)$,
$[3,6)$ and $[6,10)$~s following~\cite{li2026eap}. We exclude their negative-scenario term, since the sequences evaluated here contain no non-collision cases. This is the quantity on which models are selected, and it scores the part of the sequence that matters for collision avoidance rather
than the part that dominates \eqref{eq:ettc}.

\subsection{Datasets and training}
\label{datasets}

\textbf{DVS128-Gesture}~\cite{amir2017gesture} contains $11$ hand and arm
gestures from $29$ subjects under three lighting conditions, recorded at
$128 \times 128$, giving an embedding table of $2 \times 128 \times 128$ rows.
We use the official split. Training windows are contiguous slices of
$131{,}072$ events. Augmentation follows the set used
by~\cite{schoneScalableEventbyeventProcessing2024}: time skew, per-event
temporal and spatial jitter, and random event dropping, together with the
geometric augmentations of~\cite{li2022nda} (roll, rotation, scaling) and event
CutMix~\cite{yun2019cutmix}. Horizontal flipping is excluded, since mirroring
exchanges gesture classes.

\textbf{EvTTC}~\cite{sun2025evttc} contains collision scenarios with cars and
pedestrian dummies recorded with Prophesee cameras at $1280 \times 720$, with
ground-truth TTC derived from synchronised GNSS and LiDAR. Unlike prior work on
this benchmark, we supply no target bounding box: REACT predicts from the full
field of view. Event coordinates are downsampled by a factor of $8$ onto a
$160 \times 90$ grid, with a $10$~ms refractory period per coarse pixel and
polarity to suppress duplicates; the embedding table therefore contains
$2 \times 90 \times 160$ rows, giving $3.7$~M parameters in total. Training
windows contain $32{,}768$ tokens, obtained by decimating a contiguous span by a
factor of $16$, roughly $0.8$~s of stream. Augmentation comprises horizontal
flipping, per-event spatial jitter, random event dropping, injected clutter
events, and a time skew that rescales the stream, which simulates a different approach speed. We additionally evaluate zero-shot transfer on
FCWD~\cite{li2024eventaided} without fine-tuning.

Models are implemented in SpikingJelly~\cite{fang2023spikingjelly} and trained
with AdamW, with weight decay disabled on the state-space parameters. We report
three seeds on DVS128-Gesture and five on EvTTC.

\section{Experiments}
\label{experiments}

\subsection{Gesture recognition}
\label{sec:dvs-gest}

We first evaluate the architecture on a standard neuromorphic classification benchmark, before turning to the continuous temporal prediction task of Sec.~\ref{sec:ttc}. DVS128-Gesture is the benchmark on which Event-SSM and its compute-in-memory counterpart both report results, so it measures whether a fully spiking SSM reaches the accuracy of comparable event-by-event models at a much smaller cost. Table~\ref{tab:dvs-gesture} compares the three. Event-SSM~\cite{schoneScalableEventbyeventProcessing2024} and the compute-in-memory implementation~\cite{zhangComputeinmemoryImplementationState2026}
achieve accuracies of $97.7\%$ and $97.3\%$, respectively, but propagate real-valued activations between layers, requiring non-spiking computation and therefore limiting their direct deployment on spike-based neuromorphic hardware. REACT achieves a comparable accuracy of $97.6\%$ while using $26\%$
fewer parameters for the SSM blocks and maintaining fully spiking computation throughout the network.

\begin{table*}
\renewcommand{\arraystretch}{1.25}
\caption{\textbf{DVS128-Gesture.} Parameter counts are reported as
embedding $+$ state-space, following~\cite{schoneScalableEventbyeventProcessing2024};
the embedding is fixed by the sensor and identical across all three models.
Activity denotes the fraction of units active per update.
$^{\ast}$Energy is estimated using the proxy described in
Sec.~\ref{res_energy}, applied to six dense real-valued SSM layers.
Best results are in \textbf{bold}; second-best results are \underline{underlined}.
}
\label{tab:dvs-gesture}
\centering
\footnotesize
\begin{tabular*}{\textwidth}{@{\extracolsep{\fill}}lccccccc@{}}
\toprule
Method & Spiking & Params (M) & Prec. & Memory (MB) & Energy (mJ) & Activity (\%) & Acc.\ (\%) \\
\midrule
Event-SSM~\cite{schoneScalableEventbyeventProcessing2024}
  & \ding{55} & $4.19 + 0.80$ & FP32 & 20.00 & $158.8^{\ast}$ & 100 & \textbf{97.7} \\
CIM-SSM~\cite{zhangComputeinmemoryImplementationState2026}
  & \ding{55} & $4.19 + 0.81$ & INT8 & \underline{5.00} & $\underline{23.7}^{\ast}$ & 100 & 97.3 \\
\midrule
\textbf{REACT (ours)}
  & \ding{51} & $4.19 + \mathbf{0.04}$ & FP32 & 16.14 & $50.6$ & \underline{31.1} & \underline{97.6} \\
\textbf{REACT (ours)}
  & \ding{51} & $4.19 + \mathbf{0.04}$ & INT8 & \textbf{4.04} & $\mathbf{7.7}$ & \textbf{31.1} & 97.2 \\
\bottomrule
\end{tabular*}
\\[2pt]
\begin{minipage}{\textwidth}\footnotesize
Energy is calculated from   $131{,}072$ events for REACT
\end{minipage}
\end{table*}

\subsection{Time-to-contact}
\label{sec:ttc}

The asynchronous nature of REACT is particularly relevant for tasks that require continuous feedback from a rapidly evolving sensory stream. We therefore apply the model to event-based time-to-contact (TTC) estimation, which provides a direct test of its ability to continuously update a temporal prediction from incoming events. Most existing approaches to event-based TTC estimation first aggregate events into pseudo-frames or voxel grids before performing the prediction. While these representations can achieve good estimation accuracy, they introduce an additional temporal latency: events
must first be accumulated over a predefined temporal interval before the network can process them.
Table~\ref{tab:ttc-related-work} compares REACT with published results on
EvTTC. Every prior entry receives the target as a tracked or ground-truth
bounding box, and Garl-TTC~\cite{li2026eap} additionally uses frames alongside
events; REACT predicts from the raw stream over the full field of view. Under
that handicap it reaches $9.59 \pm 0.74\%$, within $0.15$ percentage points of
the best learned method. Parameter counts are not directly comparable across
rows: REACT's are dominated by the per-pixel embedding table, which replaces a
convolutional front-end rather than supplementing it. The frame-based
focus-of-expansion baseline~\cite{stabinger2016monocular} reaches a lower error
still, but operates offline on rectified frames with the target given and
reports no inference time, so it does not constrain a real-time budget.

\begin{figure}
    \centering
    \includegraphics[width=\columnwidth]{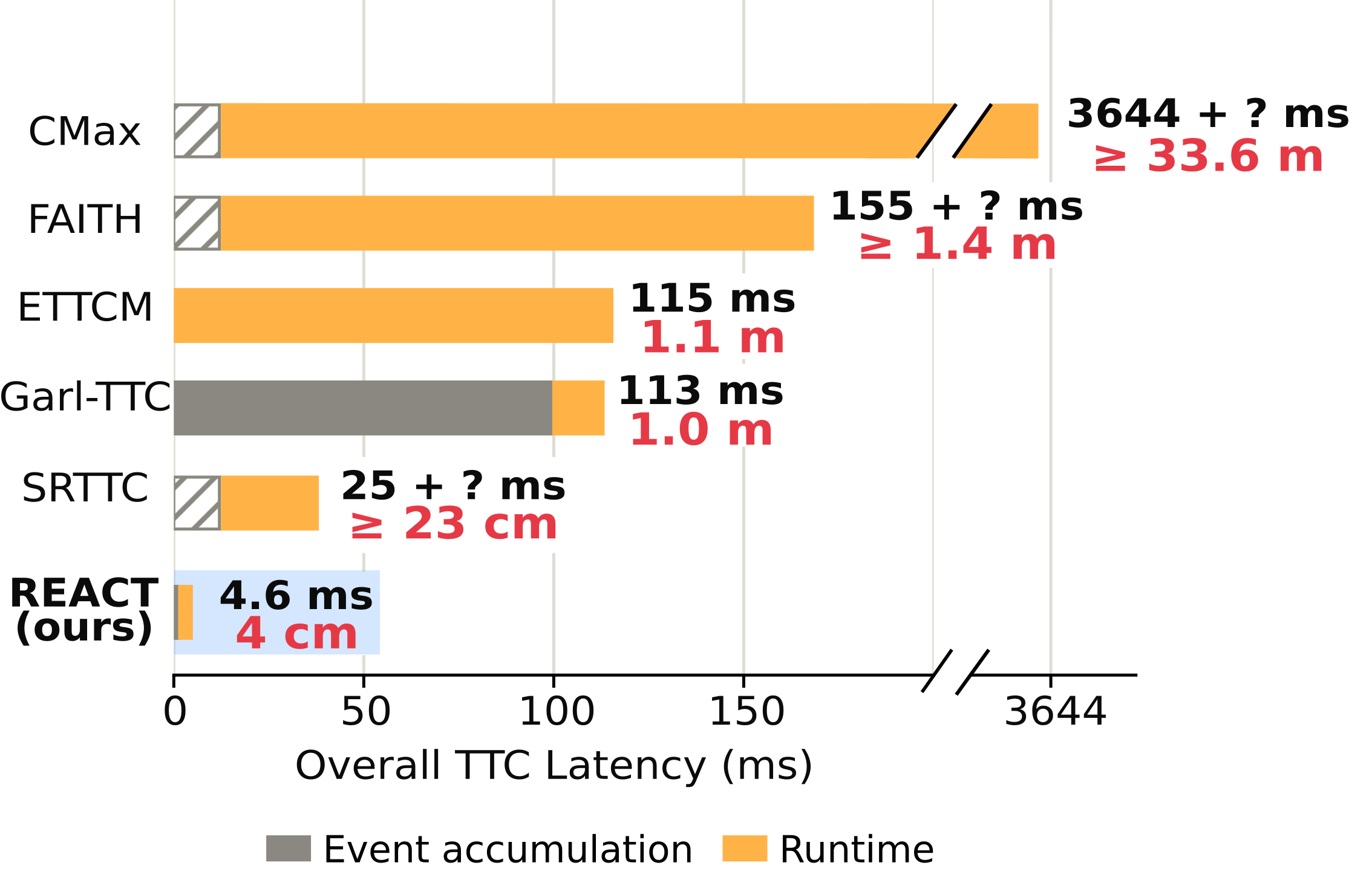}
\caption{\textbf{Latency breakdown of time-to-contact (TTC) methods},
including event accumulation (grey) and runtime (orange). Diagonal hatching
indicates methods for which the event acquisition time is not reported; in
these cases, only a lower bound on the overall latency can be provided.
Red annotations indicate the corresponding distance travelled during this
latency, computed using the mean vehicle speed near the target in the EvTTC
dataset~\cite{sun2025evttc} (9.22\,m/s).}
    \label{fig:latency}
\end{figure}

Accumulation latency becomes critical when the platform moves rapidly. Conventional approaches must integrate events over a finite temporal interval before producing a prediction, during which the vehicle continues to move. This creates an unavoidable gap between when visual information is acquired and when the corresponding estimate becomes available. Event-by-event processing avoids this integration delay: REACT updates its internal state and TTC estimate as events arrive, so the sensing-to-prediction latency is determined primarily by computation. Figure~\ref{fig:latency} quantifies this difference. For frame- and window-based methods, overall latency is dominated by the integration window rather than network inference, resulting in the totals reported in Table~\ref{tab:ttc-related-work}: $113$~ms for Garl-TTC~\cite{li2026eap}, $115$~ms for ETTCM~\cite{nunes2023ttcmap}, and at least $3644$~ms for CMax~\cite{gallego2018unifying}. In contrast, REACT produces an updated estimate with an end-to-end latency of only $4.6$~ms. At the mean approach speed of $9.22$~m/s measured near the target in EvTTC, the vehicle travels only $4$~cm during this interval, compared with more than $1$~m for the fastest of the window-based alternatives. 

\begin{figure}[t]
    \centering
    \includegraphics[width=\columnwidth]{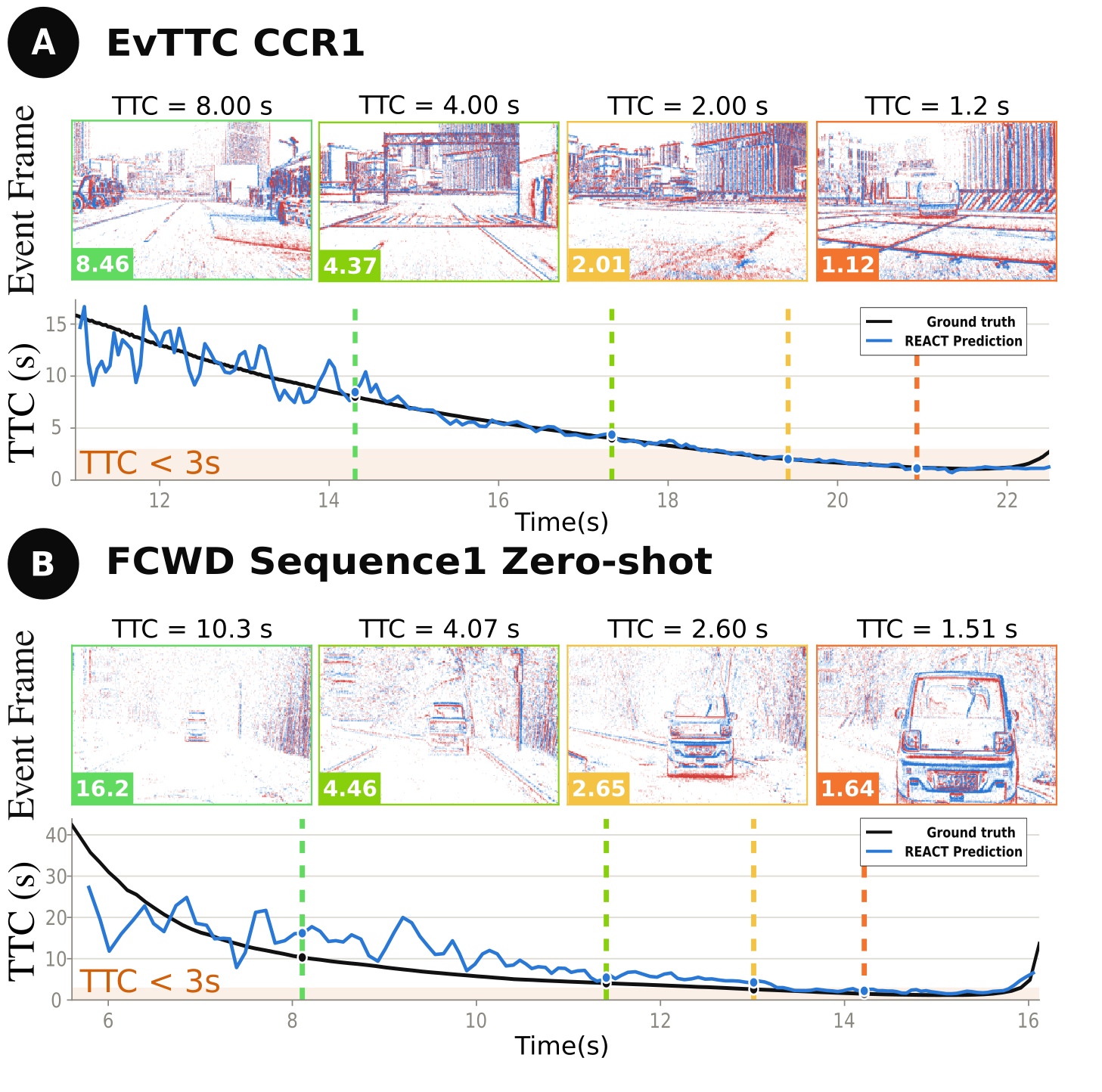}
    \caption{\textbf{Qualitative TTC predictions over complete sequences.}
    (\textbf{A}) In-domain evaluation on the CCR1 split of the medium-overlap-50 set of EvTTC~\cite{sun2025evttc}. (\textbf{B}) Zero-shot evaluation on Sequence 1 of FCWD~\cite{li2024eventaided}. For each sequence, four event frames are shown at decreasing time to contact. The values above each frame indicate the ground-truth
    TTCs, while the colored labels in the lower-left corners show the corresponding REACT predictions (in seconds). Predictions progressively converge toward the ground truth as the time to contact decreases, with reliable estimates obtained
    within approximately 3\,s of impact (see the critical zone in orange). REACT exhibits similar temporal prediction behavior on the unseen FCWD sequence under the domain shift, without fine-tuning. In this cases, the average vehicle speed is 9.22 m/s.
    }
    \label{fig:quantitative}
\end{figure}

\begin{table*}
  \centering
\caption{\textbf{EvTTC time-to-collision.} All prior methods receive the target
as a tracked or ground-truth bounding box, whereas REACT predicts directly from
the raw event stream without a target prior, using the in-domain validation
split of~\cite{sun2025evttc}. Inference time denotes the end-to-end latency
reported in Fig.~\ref{fig:latency}; $\geq$ indicates methods that do not report
an accumulation window. $^{\ast}$Energy is estimated using the proxy described
in Sec.~\ref{res_energy}, applied to Garl-TTC's ResNet-50 backbone; no prior
method reports energy. Best results per column are in \textbf{bold}.}
  \label{tab:ttc-related-work}
  \resizebox{\textwidth}{!}{%
    \begin{tabular}{@{}lllccccccc@{}}
      \toprule
      Method & Category & Modality & Params & Prec. & $e_{\mathrm{TTC}}$ (\%) & Target-free & Async. & Time (ms) & Energy (mJ) \\
      \midrule
      Image's FoE~\cite{stabinger2016monocular} & Geometric & Frame & - & - & 3.51 & \ding{55} & \ding{55} & - & - \\
      CMax~\cite{gallego2018unifying} & Geometric & Event & - & - & 9.11 & \ding{55} & \ding{55} & $\geq 3644$ & - \\
      Garl-TTC~\cite{li2026eap} & Learned & Event + Frame & $23.5 M^{\ast}$ & FP32 & 9.44 & \ding{55} & \ding{55} & $113$ & $197^{\ast}$ \\
      STRTTC~\cite{li2024eventaided} & Geometric & Event & - & - & 11.22 & \ding{55} & \ding{55} & $\geq 25$ & - \\
      FAITH~\cite{dinaux2021faith} & Geometric & Event & - & - & 40.97 & \ding{55} & \ding{55} & $\geq 155$ & - \\
      ETTCM~\cite{nunes2023ttcmap} & Geometric & Event & - & - & 50.38 & \ding{55} & \ding{55} & $115$ & - \\
      \midrule
      \textbf{REACT (ours)} & Learned & Event & $\mathbf{3,7 M}$ & FP32 & $9.59 \pm 0.74$ & \ding{51} & \ding{51} & $\mathbf{4.6}$ & $18.5$ \\
      \textbf{REACT (ours)} & Learned & Event & $\mathbf{3,7 M}$ & \textbf{INT8} & $10.26 \pm 0.81$ & \ding{51} & \ding{51} & $\mathbf{4.6}$ & $\mathbf{2.8}$ \\
      \bottomrule
    \end{tabular}%
  }
  \\[2pt]
  \begin{minipage}{\textwidth}\footnotesize
  Energy is calculated from $32{,}768$ events for REACT. \\
  $^{\ast}$Estimated parameter count with a ResNet-50.
  \end{minipage}
\end{table*}

\subsubsection{Zero-shot transfer}
To test whether the learned temporal representation is specific to EvTTC, we
evaluate the same trained model on FCWD~\cite{li2024eventaided} without
fine-tuning. FCWD was recorded with a different sensor and under different
driving conditions, so the transfer is a test of the representation rather than
of the training distribution. Table~\ref{tab:zeroshot} reports the resulting
errors alongside the in-domain figures, and Fig.~\ref{fig:quantitative} shows
complete predicted trajectories in both settings. The behaviour is consistent
across the domain shift: predictions are unreliable while the target is
distant, where the expansion signal in the event stream is weak, and converge as
the collision approaches. The error therefore concentrates in the region that
\eqref{eq:ettc} weights most heavily and that matters least operationally, which
is what motivates $\mathrm{MiD}_w$ (Sec.~\ref{sec:task}) as the selection
criterion.

\begin{table}[t]
\renewcommand{\arraystretch}{1.25}
\caption{\textbf{Zero-shot transfer to FCWD}~\cite{li2024eventaided}.
The same model trained on EvTTC is evaluated on FCWD without fine-tuning or
domain adaptation.}
\label{tab:zeroshot}
\centering
\footnotesize
\begin{tabular}{@{}lcc@{}}
\toprule
Evaluation & $e_{\mathrm{TTC}} \downarrow $ (\%) & $\mathrm{MiD}_w \downarrow$ \\
\midrule
EvTTC (in-domain)   & $9.322$ & $0.0986$ \\
FCWD (zero-shot)    & $25.7$            & $0.246$ \\
\bottomrule
\end{tabular}
\\[2pt]
\begin{minipage}{\columnwidth}
\end{minipage}
\end{table}

\subsection{Anytime accuracy}
\label{sec:anytime}

Because no accumulation window is required, REACT can be queried at any instant,
including instants at which it has seen very little of the stream. To measure
how much stream it needs, we empty the state of a trained checkpoint
$\delta$~ms before a query, let the model consume only the events of that
interval, and record the median error at the query.

An estimate is usable long before a frame-based pipeline could assemble its
input: $27\%$ median error from $100$~ms of stream, which is the earliest
instant at which a two-frame method such as Garl-TTC~\cite{li2026eap} has its
second frame, and $20\%$ from $200$~ms. Accuracy peaks at $12.9\%$ around
$400$~ms and then degrades to $21\%$. By not having any form of reset in the internal state and the readout of the model, its predictions is refined as times goes on, making REACT suitable for instant prediction and high-horizon predictions (see Figure~\ref{fig:quantitative} to see model stabilization of prediction in function of the time)

\subsection{Network quantization}
\label{quantization}

To implement SNNs on dedicated neuromorphic hardware, the network needs to be quantized (usually on from full-precision 32-bits (FP32) to integer 8 bits (INT8). 
Following a similar methodology as in \cite{zhangComputeinmemoryImplementationState2026}, we use quantization-aware training. From a converged FP32 model we fine-tune the model for $150$ epochs by converting only the synaptic weights and the neuron parameters to INT8 as the activation functions are already binary. 
By clamping the weights in INT8 and making the update and the full precision weights. This methods allows the model to fine tune progressively using full-precision copy for weight update.

Over three different seeds, this gives $97.2 \pm 0.4\%$ against $97.6 \pm 0.3\%$ in FP32 on DVS-Gesture dataset.

\subsection{Ablation Study}
\label{res_ablation}

\subsubsection{TTC Full-Frame}
Every published method on EvTTC receives the target as a bounding box, either provided by a detector and tracker or obtained from ground truth. This removes distractors from the field of view and reduces TTC estimation to the temporal evolution of a known target region~\cite{li2026eap}. While this setup is standard and effective, it introduces a separate localization stage before TTC estimation and therefore does not provide a fully asynchronous end-to-end pipeline. To disentangle the contributions of target localization and temporal estimation, we evaluate the same REACT configuration both on the full field of view and on target crops. On the held-out sequences, the TTC error decreases from 34.2\% with the full frame to 28.3\% with target crops, indicating that a substantial part of the gap between REACT and published results stems from target localization rather than temporal estimation. This also suggests that incorporating an ROI front-end could improve performance without modifying the underlying spiking temporal model.

\subsubsection{Neuron model}
The temporal dynamics of REACT strongly rely on its neuron model, making
neuron replacement the natural ablation while keeping the rest of the
architecture fixed. We compare C-SiLIF against SiLIF~\cite{fabreSiLIFStructuredState2026},
which retains the log-reparameterized decay and learnable timestep but
remains real-valued and therefore lacks an oscillatory mode, and against a
standard LIF neuron, whose dynamics are governed by a single fixed decay
constant. All variants are trained on EvTTC with the same two residual
blocks, $D=128$ channels, event budget, and training schedule. We evaluate
LIF both in its standard form, with a spike-triggered reset, and with the
reset removed as described in Sec.~\ref{sec:csilif}. This isolates the
contribution of the linear recurrence from that of the neuron dynamics
themselves.

\begin{table}[t]
\renewcommand{\arraystretch}{1.25}
\caption{Neuron model on EvTTC. All variants share the architecture, event
budget and training schedule of Sec.~\ref{sec:arch}; only the neuron differs.
Best in \textbf{bold}, second \underline{underlined}.}
\label{tab:neuron}
\centering
\footnotesize
\begin{tabular}{@{}lccc@{}}
\toprule
Neuron & State & $e_{\mathrm{TTC}}$ (\%) $\downarrow$ & $\mathrm{MiD}_w$ $\downarrow$ \\
\midrule
LIF                                        & real, fixed decay, reset & $116.9$ & $0.681$ \\
LIF, no reset                              & real, fixed decay        & $84.1$  & $0.52$ \\
SiLIF~\cite{fabreSiLIFStructuredState2026} & real, learned decay      & \underline{$11.42$} & \underline{$0.12$} \\
\textbf{C-SiLIF}~\cite{fabreSiLIFStructuredState2026} & complex, oscillatory & $\mathbf{9.3}$ & $\mathbf{0.09}$ \\
\bottomrule
\end{tabular}
\end{table}

The results are reported in Table~\ref{tab:neuron}. Removing the reset alone
reduces the $e_{TTC}$ from $116.9\%$ to $84.1\%$, showing that preserving the
state across spikes is already critical for temporal prediction. With a
spike-triggered reset, the neuron repeatedly discards its accumulated temporal
state, severely limiting its ability to represent long-range temporal
dependencies. Both LIF variants nevertheless remain far from usable in our
experimental setting, highlighting the importance of the SSM-like temporal
dynamics. Replacing LIF with SiLIF reduces the error to $11.42\%$, while
C-SiLIF further reduces it to the reported $9.3\%$ on EvTTC. The distance traveled during the latency, and thus the effective blind distance, increases linearly with vehicle speed. At 100 km/h, these latencies correspond to only 12.8 cm of blind travel for REACT, compared with 3.14 m for Garl-TTC and 3.19 m for ETTCM, highlighting the practical advantage of our approach.

\subsection{Energy consumption}
\label{res_energy}

We estimate inference energy for a neuromorphic deployment using the 45\,nm energy proxy proposed in~\cite{lemaireAnalyticalEstimationSpiking2023}. For
each layer, the estimated energy is computed as $(\text{reads}+\text{writes}) E_{\text{SRAM}} +\text{acc}\,E_{\text{add}}$, where the number of synaptic operations is determined by the number of input spikes, $\theta_{\text{in}}$, measured on
real data. We additionally account for two costs specific to our
architecture. First, the event embedding requires $D$ word reads per input event. Second, the C-SiLIF core reads five per-channel parameters ($\lambda_{\mathrm{re}},\lambda_{\mathrm{im}},\rho,b_{\mathrm{re}},
b_{\mathrm{im}}$) at each update, compared with a single decay parameter for a conventional LIF neuron, and requires $12$ multiplications and $5$ additions per event.

For energy estimation, we consider the sequential deployment of the model, since the parallel associative scan described in Sec.~\ref{methods} is used only for training and batched sequence processing.

Firing rates are measured independently for each block. On DVS128-Gesture, the overall firing rate is $31.1 \pm 1.3\%$ across three seeds, with $31.9 \pm 1.3\%$ and $6.3 \pm 0.4\%$ in blocks~0 and~1, respectively. On EvTTC, the overall firing rate is $23.2 \pm 1.2\%$ across five seeds. INT8 quantization does not alter the binary spike activity and therefore
leaves the firing rates unchanged. We consequently use the same measured rates for the FP32 and INT8 energy estimates, with the difference in energy arising solely from the reduced per-operation cost.

The resulting estimated energy per input window is reported in
Tables~\ref{tab:dvs-gesture} and~\ref{tab:ttc-related-work}. Channel-mixing linear layers account for approximately $90\%$ of the total estimated energy, while the C-SiLIF state-space core represents the lowest-cost major component. The estimated energy per event is $2.7\times$ higher for EvTTC than for DVS-Gesture, primarily because event pooling is not used for EvTTC. For EvTTC, where the input windows are bounded in time, the estimated energy scales linearly with the number of events at a fixed firing rate, yielding an estimated cost of $563\,\mathrm{nJ}$ per event.

\section{Discussion}
\label{discussion}

We introduced REACT, a new fully spiking state-space model that processes
raw event streams event by event, without frames, voxel grids, or fixed
temporal bins. By combining binary neural activity with continuous-time
state-space dynamics driven by physical inter-event intervals, REACT
retains temporal information while avoiding explicit temporal
discretization. On DVS128-Gesture, it achieves competitive accuracy
with event-by-event non-spiking SSMs while using a substantially smaller
state-space component, demonstrating that rich temporal dynamics can
be implemented with a compact spiking formulation.

The TTC experiments illustrate the relevance of this approach for
robotic perception. REACT achieves $9.59\pm0.74\%$ error on EvTTC,
comparable to the best reported learning-based method, while requiring
no target bounding box as input. More importantly, processing events
individually yields a 4.6~ms end-to-end inference latency, avoiding the
accumulation delays introduced by frame- or window-based approaches.
The zero-shot experiment further indicates that the learned temporal
representation can transfer across datasets, although performance
degrades for distant targets.

The energy analysis also provides insight into the architecture.
INT8 quantization causes only a limited accuracy degradation and reduces
the estimated EvTTC energy to 2.8~mJ per 32,768 events. However,
approximately 90\% of the estimated energy is spent in channel-mixing
layers rather than in the C-SiLIF state-space core. This suggests that
further sparsification of these operations may offer greater efficiency
gains than optimizing the temporal dynamics themselves. These values
remain analytical estimates and should therefore be validated on
neuromorphic hardware.

Finally, the embedding remains a major source of parameters and scales
with sensor resolution. Factorized or coordinate-based embeddings, along
with training on more diverse motion scenarios, are promising directions
for improving scalability and robustness. Evaluation on physical
neuromorphic hardware and robotic platforms will be an important step
toward deployment.

\section{Conclusion}
\label{conclusion}

REACT places the dynamics of a state-space model inside the spiking neuron. Processing events asynchronously brings learned models closer to anytime prediction, where an estimate is available whenever it is needed rather than once a window has closed. Nothing in the architecture is specific to
time-to-contact: any task that maps a continuous event stream to a continuously updated estimate, such as steering-angle prediction, ego-motion estimation or place recognition, would be a natural fit for future ongoing works. Ultimately, such event-driven anytime perception could provide the low-latency feedback required to connect asynchronous vision directly to closed-loop robotic control.

\section*{Acknowledgments}
We acknowledge the use of ChatGPT and Claude for assistance with
language editing, clarity, and phrasing throughout the manuscript. Claude was also used for coding assistance, including Triton implementations for accelerated GPU training, factorization, and plot generation.

\bibliographystyle{IEEEtran}
\bibliography{references}

\end{document}